\documentclass[10pt,twocolumn,letterpaper]{article}

\usepackage{iccv}
\usepackage{times}
\usepackage{epsfig}
\usepackage{graphicx}
\usepackage{amsmath}
\usepackage{amssymb}
\usepackage{multirow}  
\usepackage[noend]{algpseudocode}
\usepackage{algorithmicx,algorithm}
\usepackage[T1]{fontenc}
\usepackage[numbers]{natbib}
\usepackage{booktabs}
\usepackage{url}

\usepackage{graphicx}

\usepackage{subfigure}
\usepackage{caption}

\newcommand{\vw}{{\bf w}}

\newcommand{\vg}{{\bf g}}

\newcommand{\reals}{{\mathbb{R}}}

\usepackage[breaklinks=true,bookmarks=false]{hyperref}

\iccvfinalcopy 

\begin{document}

\title{ Low Dimensional Landscape Hypothesis is True: \\DNNs can be Trained in Tiny Subspaces}

\author{Tao Li\textsuperscript{1}
\and
Lei Tan\textsuperscript{1}
\and
Qinghua Tao\textsuperscript{2}
\and
Yipeng Liu\textsuperscript{3}
\and
Xiaolin Huang*\textsuperscript{1}
\and
\textsuperscript{1}Shanghai Jiao Tong University
\and
\textsuperscript{2}Tsinghua University
\and
\textsuperscript{3}University of Electronic Science and Technology
}
\maketitle
\ificcvfinal\thispagestyle{empty}\fi

\begin{abstract}
Deep neural networks (DNNs) usually contain massive parameters, but there is redundancy such that it is guessed that the DNNs could be trained in low-dimensional subspaces. In this paper, we propose a Dynamic Linear Dimensionality Reduction (DLDR) based on low-dimensional properties of the training trajectory. The reduction is efficient, which is supported by comprehensive experiments: optimization in 40 dimensional spaces can achieve comparable performance as regular training over thousands or even millions of parameters. Since there are only a few optimization variables, we develop a quasi-Newton-based algorithm and also obtain robustness against label noises, which are two follow-up experiments to show the advantages of finding low-dimensional subspaces.
\end{abstract}

\section{Introduction}

Deep neural networks (DNNs) have achieved unprecedented success in various fields \cite{lecun1998gradient,lecun2015deep}. In DNNs, the number of parameters is usually very large, e.g., 28.5M in VGG11 \cite{simonyan2014very}, 3.3M in MobileNet \cite{howard2017mobilenets}, and 21.0M in Xception \cite{chollet2017xception}.
However, simply regarding each parameter of DNNs as an independent variable is too rough. In fact, the parameters have strong mutual relationships. For example, the gradient is propagated from deep layers to shallow ones and hence the gradients of parameters are strongly related \cite{hecht-nielsen1989theory}.
The parameters in the same layers also have synergy correlations.
Therefore, the number of \emph{independent} optimization variables may be not as many as the number of parameters. In other words, it comes a hypothesis that
the landscape of the DNNs' objective functions can be in a relatively low-dimensional subspace, which is first raised by \cite{gur2018gradient}.





If the low-dimensional landscape hypothesis is true, DNNs' training could be conducted in low-dimensional landscapes and there will be great benefits in  learning from both practical and theoretical aspects.
The golden criterion is whether optimization in such low-dimensional spaces could achieve the same or similar performance as optimizing all parameters in the original space. In the pioneering work \cite{li2018measuring}, the authors set $90\%$ accuracy of the SGD training on full parameters as the criterion and find that the intrinsic dimension needed for training is much smaller than the number of parameters. For example, on CIFAR-10 \cite{krizhevsky2009learning}, LeNet \cite{lecun2015lenet} with 62006 parameters could be optimized in 2900-dimensional subspaces and the accuracy is $90\%$ of regular training.
Albeit that method, in which subspaces are extracted by random projections, is preliminary, the performance is very promising. Later, \cite{gressmann2020improving} considers different parts of the network and re-draws the random bases at every step. Then, the landscape dimensionality is further reduced to hundreds and accuracy downgrading is similar.



In this paper, we propose to extract the landscape through analyzing the optimization trajectory, instead of random projection in \cite{li2018measuring}, \cite{gressmann2020improving}.
Via the proposed method, 
many standard  neural network architectures could be well trained by only $40$ independent variables and the performance is almost as same as the regular training on all parameters, showing that
\begin{center}
   \emph{DNNs can be trained in low-dimensional subspaces} \\
\end{center}
and we indeed can effectively find such landscapes.

In this paper, we use $f(x,\vw)$ to denote a DNN and $\vw \in \reals^n$ to the parameters. Then the parameters' training sequence can be denoted as  $\{\vw_i\}_{i=0,\ldots,t}$, where $\vw_i$ refers to the value of $\vw$ at the training step $i$. The low-dimensional landscape hypothesis means that we can find a  subspace (actually, it is an affine  set but we will do centralization later, and hence we do not strictly distinguish these two concepts in this paper) to approximately cover the optimization trajectory $\{\vw_i\}_{i=0,\ldots,t}$, which is based on the low-rank property of Neural Tangent Kernel (NTK, \cite{jacot2018neural,lee2019wide}) and will be explained in Section \ref{sec:motivation}.

\begin{figure}[!htp]
	\centering
	\includegraphics[angle=0,origin=br,width=6.5cm]{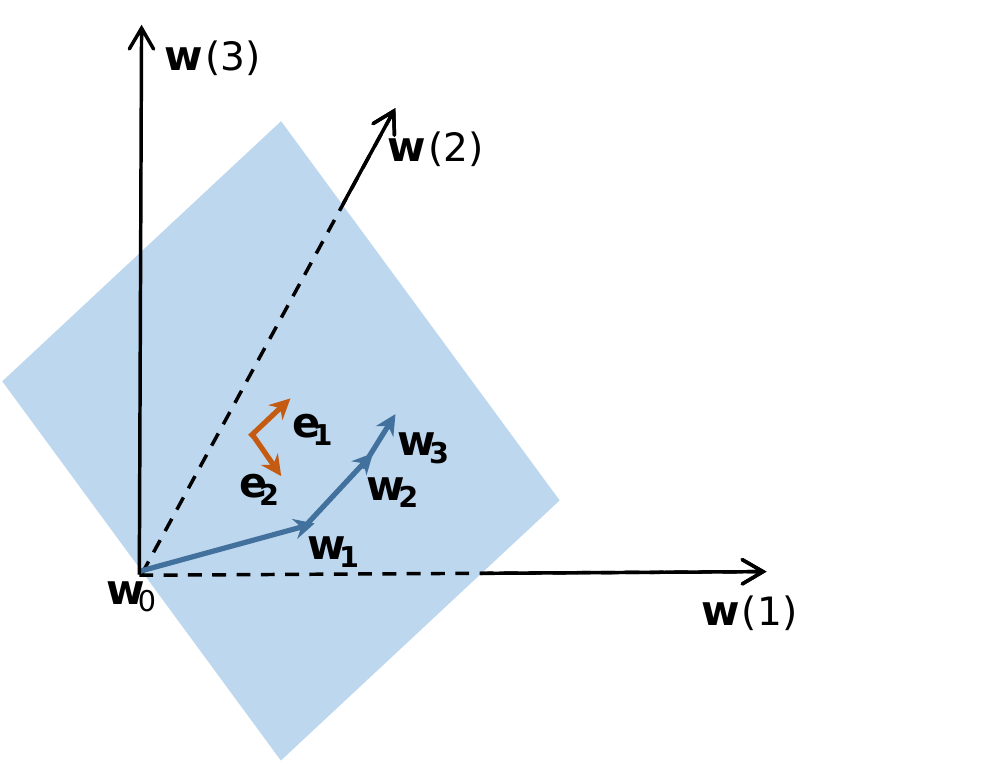}
	\caption{
	There are three parameters $\vw(1),\vw(2),\vw(3)$ to optimize. But the training trajectory  $\{\vw_i\}_{i=0,\ldots,t}$ could be in a two-dimensional subspace spanned by $\mathbf{e}_1$ and $\mathbf{e}_2$. If so, training in the low-dimensional space can have comparable performance as training in the high-dimensional space.
	}
	\label{traj_1}
\end{figure}

Notice that extracting the landscape, i.e.,
finding independent variables, is different from selecting parameters as in model reduction methods, see, e.g., \cite{srinivas2015data,frankle2018lottery}. Consider a  toy example in Fig. \ref{traj_1}, which contains three  variables to be optimized. As shown the optimization trajectory is in a subspace spanned by $\mathbf{e}_1$ and $\mathbf{e}_2$, i.e., the dimensionality of the optimization landscape is $2$, but there is no single parameter which could be reduced. This simple example shows our focus: we aim to find a suitable combination of parameters to construct independent variables which are in the low-dimensional landscapes.

To find the subspace to cover the dynamic trajectory should care about the training dynamics. Therefore, we  name the proposed subspace extracting method as Dynamic Linear Dimensionality Reduction (DLDR). With independent variables obtained from DLDR, we can well characterize the training dynamics in a subspace spanned by only a few bases (independent variables). In Section \ref{Experiments}, extensive numerical experiments will show that many standard  DNN architectures can be well trained by only $40$ independent variables and meanwhile the test accuracy can be maintained almost as the same as regular training on full parameters in the original space.



In theory, reducing millions of parameters to only a few independent variables could explain the good generalization performance of DNNs even when the training set is not very big. In practice, as only a few independent variables are needed to be optimized, we can apply second-order methods rather than first-order optimization methods, e.g., SGD \cite{ruder2016overview}, to overcome some inherent drawbacks, such as scale-sensitivity and slow convergence. In the existing works, part of second-order information like momentum has been introduced and has significantly improved DNNs' performance, leading to the current popular adaptive strategies like
Adam \cite{kingma2014adam}, RMSprop \cite{dauphin2015rmsprop},  etc. Thanks to the low-dimensional subspace found by DLDR, quasi-Newton methods, such as DFP and BFGS \cite{broyden1970convergence, byrd1994representations}, become applicable to the training of DNNs. In this paper, a BFGS algorithm on the projected landscape, hence called \emph{P-BFGS}, is proposed and obtains about $30\%$ time-saving from SGD.

Another follow-up application of the low-dimensional hypothesis is to improve the robustness against label noises. Since DNNs are working the over-parameterization scenery, they could easily fit any label, even incorrect and meaningless ones \cite{zhang2021understanding}. Therefore,
when the trainning labels are corrupted by noises, the DNNs could be easily destroyed. Now, we have found very low dimensional landscapes, and training DNNs in such landscapes is expected to be more robust to label noises. In Section \ref{LabelNoise_sec}, we will find that without any other robustness enhancement techniques, training in low-dimensional spaces attains over $50\%$ test accuracy on CIFAR-10, even when $90\%$ of the training labels are set randomly.

The most important contribution of this paper is to verify the low-dimensional landscape hypothesis: DNNs'  training trajectory can be covered by a very low-dimensional subspace, and optimization in such a low-dimensional subspace could achieve similar performance as training all parameters. Detailed contributions include,
\begin{itemize}
    \item 
    We design a Dynamic Linear Dimensionality Reduction technique to efficiently find the low-dimensional space. Compared with the existing methods, the accuracy of DLDR is significantly improved and the dimensionality gets  an order of magnitude decrease.

    \item 
    We develop a quasi-Newton-based algorithm, i.e., P-BFGS, which brings efficiency in training and shows the possibility of using second-order methods for training DNNs.

    \item 
    We conduct experiments on label noises, showing the inherent robustness of training in low-dimensional subspaces. 
\end{itemize}

The rest of this paper is organized as follows. We first review the related works in Section \ref{related works}. Then  DLDR algorithm for low-dimensional landscape hypothesis is proposed and verified in Section \ref{DLDRsec}. Based on  DLDR, we design a quasi-Newton algorithm in Section \ref{BFGSsec}. We then evaluate dimensionality reduction performance in
Section \ref{Experiments}.
Section \ref{discussion} ends the paper with a brief discussion.
The code is released$\footnote{\url{https://github.com/nblt/DLDR}}$.

\section{Related Works}
\label{related works}

Analyzing and understanding the landscape of DNNs' optimization objective is of great importance.
For example, Li et al. \cite{li2018visualizing} visualize the loss landscape of DNNs using a range of visualization methods. He et al. \cite{he2019asymmetric} observe that there are many asymmetric directions at a local optimum along which the loss increased sharply on one side and slowly on the other.
One important aspect is to measure the
intrinsic dimensionality of DNNs' landscape. In the pioneering work \cite{li2018measuring}, it is found that with random projection, optimization in a reduced subspace can reach $90\%$ performance of regular SGD training. Based on that it is claimed that the intrinsic dimensionality is much smaller than the number of parameters.
The following work \cite{gressmann2020improving} improves the random bases training performance by
considering different parts of the network and re-drawing the random bases at every step. Different from previous works, we extract the subspace via analyzing the DNNs' training dynamics and then get significant improvement: the intrinsic dimensionality is reduced by a order of magnitude and the accuracy is improved to be almost the same as regular training.

Verifying the low-dimensional hypothesis and finding the tiny subspace in which DNNs could be well trained are very important, not only for theoretical discussion but also for practical learning. In theory, it coincides with the discovery in  \cite{gur2018gradient} that after a short period of training, the gradients of DNNs can converge to a very small subspace spanned by a few top eigenvectors of the Hessian matrix. In practice, the low-dimensional hypothesis may inspire the design of more powerful optimization methods and bring more potentials to overcome some existing barriers in learning.
Since in the tiny subspace, the number of optimization variables is largely reduced and hence higher-order information can be utilized in a relatively easy way. In \cite{tuddenham2020quasi}, \cite{sohl2014fast}, etc, delicate methods are designed to use curvature information while keeping computation efficiency, which is however an incompatible contradiction unless the number of optimization variables can be effectively reduced.


\section{Dynamic Linear Dimensionality Reduction} \label{DLDRsec}

\subsection{Low-rank Approximation}\label{sec:motivation}

To investigate the training trajectory and the low-dimensional landscape, we formulate the gradient flow of a single-output neural network, i.e.,
\begin{align}
\dot \vw_t = -  \nabla_{\vw} f(\mathcal{X},\vw_t)^{\top} \nabla_{f_t(\mathcal{X},\vw_t)}\mathcal{L},
\end{align}
where  $\mathcal{X}$ stands for the training set of size $m$, $\mathcal{L}$ refers to the loss function, and $\nabla_{\vw} f(\mathcal{X},\vw_t)\in \mathbb{R} ^{m \times n}$ denotes the gradients.
As shown in \cite{lee2019wide}, with infinite-width limit, a wide neural network estimator can be approximated by a linearized model under gradient descent, such that 
\begin{equation}
    \label{linearizedmodel}
    \begin{aligned}
    f^{\mathrm{lin}}(x,\vw_t) \approx f(x,\vw_0) + \nabla_{\vw} f(\mathcal{X},\vw_0)(\vw_{t}-\vw_0),
    \end{aligned}
\end{equation}
where $f^{\mathrm{lin}}$ represents the linearized model. Thus, the dynamics of gradient flow using this linearized function are governed by
\begin{equation}
\label{dynamic1}
\dot \vw_t = -  \nabla_{\vw} f(\mathcal{X},\vw_0)^{\top} \nabla_{f^{\mathrm{lin}}(\mathcal{X},\vw_t)}\mathcal{L}.
\end{equation}
In other words, the parameter dynamics are governed by $\nabla_{\vw} f(\mathcal{X},\vw_0)$ and $\nabla_{f^{\mathrm{lin}}(\mathcal{X},\vw_t)}\mathcal{L}$, the former of which is a constant matrix and the latter changes over time $t$.

Seemingly, the parameters $\vw$ vary in an $n$-dimensional  space. However, under infinite-width setting and according to (\ref{dynamic1}), the  trajectory $\{\vw_i\}_{i=0,\ldots,t}$ depends on $\nabla_{\vw} f(\mathcal{X},\vw_0)$. Here the key observation is that if $\nabla_{\vw} f(\mathcal{X},\vw_0)$ can be approximated by a low-rank matrix, we can  conduct effective dimensionality reduction on the parameter space.
To clearly show this,
we apply Singular Value Decomposition (SVD) on $\nabla_{\vw} f(\mathcal{X},\vw_0)$ and have
\begin{align}
\nabla_{\vw} f(\mathcal{X},\vw_0) = U_0 \Sigma_0 V_0^{\top},
\end{align}
where $U_0\in \mathbb{R} ^{m \times m}$ and $V_0\in \mathbb{R} ^{n \times n}$ are two real orthogonal matrices, and $\Sigma_0\in \mathbb{R} ^{m \times n}$ is a semi-definite diagonal matrix with its diagonals  $\{\lambda_i\}_{i=1,\ldots,m}$ being the singular values of $\nabla_{\vw} f(\mathcal{X},\vw_0)$ in the decreasing order. Recalling the definition of NTK, it can be rewritten as,
\begin{align}
\Theta_0 =\nabla_{\vw} f(\mathcal{X},\vw_0)\nabla_{\vw} f(\mathcal{X},\vw_0)^{\top} = U_0 \Sigma_0\Sigma_0^{\top} U_0^{\top}.
\end{align}
This formula actually represents  the spectral decomposition of $\Theta_0$. Denote $\Sigma^{\mathrm{NTK}}=\Sigma_0\Sigma_0^{\top}\in\mathbb{R} ^{m \times m}$ as a diagonal matrix with its diagonals $\{\lambda^{\mathrm{NTK}}_i\}_{i=1,\ldots,m}$ being the eigenvalues of $\Theta_0$ in decreasing order. We can obtain that $\forall i\in \{1,\ldots,m\},\lambda^{\mathrm{NTK}}_i = \lambda_i^2$. According to the recent work of Fan \textit{et al.} \cite{fan2020spectra}, under the infinite-width assumption, the spectral decomposition of the NTK converges to a certain probability distribution with Stieltjes transform. Such probability distribution empirically indicates the eigenvalue decay of the NTK: only a very small part of $\{\lambda^{\mathrm{NTK}}_i\}_{i=1,\ldots,m}$ are dominant eigenvalues in $\Theta_0$. Thus, $\{\lambda_i\}_{i=1,\ldots,m}$ enjoys a similar decay. 
Then 
$\Sigma_0$ could be approximated by a low-rank diagonal matrix denoted by $\tilde \Sigma_0 \in \reals^{d \times d}$, which contains the first $d$ largest singular values of $\Sigma_0$, i.e.,
\begin{align}
\Sigma_0 \approx \tilde U_0 \tilde\Sigma_0 \tilde V_0^{\top},
\end{align}
with $ \tilde U_0\in\mathbb{R} ^{m \times d}$ and $ \tilde V_0\in\mathbb{R} ^{n \times d}$.
Thus, $\nabla_{\vw} f(\mathcal{X},\vw_0)$ and $\dot\vw_t$  can be approximated by low-rank matrices, such that
\begin{equation}
\nabla_{\vw} f(\mathcal{X},\vw_0) \approx U_0\tilde U_0 \tilde\Sigma_0 \tilde V_0^{\top}V_0^{\top},
\end{equation}
\begin{equation}
\label{dynamic3}
\dot \vw_t \approx -\underbrace{V_0 \tilde V_0}_{\text{Variables}} \underbrace{( \tilde\Sigma_0 \tilde U_0^{\top} U_0^{\top} \nabla_{f^{\mathrm{lin}}(\mathcal{X},\vw_t)}\mathcal{L})}_{\text{Projected Gradient}}.
\end{equation}
In fact, the parameter trajectory $\{\vw_i\}_{i=0,\ldots,t}$ and the parameter evolution $\dot \vw_t$ fall into two similar spaces, between which the only difference is an affine translation.
In this regard, (\ref{dynamic3}) indicates that $\{\vw_i\}_{i=0,\ldots,t}$ and $\dot \vw_t$ could both be well embedded in a $d$-dimensional space.
$V_0 \tilde V_0$, of which the columns represent the independent variables, actually establishes a projection from the $d$-dimensional variable subspace back to the $n$-dimensional parameter space, and the second part in the right-side term of (\ref{dynamic3}) plays a role of gradient in the projected low-dimensional subspace.

The theoretical discussion implies that for neural networks training, we can get good performance using the projected gradient in particular subspaces.
However, there are pre-conditions for the above discussions: i) the width is unlimited; ii) the training is in the lazy regime.
Of course, the above two conditions are too theoretically ideal.
In the rest of this paper, empirical studies are conducted on standard DNNs to evaluate such low-dimensional property also works in practical learning,
i.e., the optimization trajectory of DNNs in real tasks could be empirically covered in very low-dimensional subspaces.

\subsection{Methodology}
The key issue of reducing the dimensionality  is to 
find the low-dimensional subspace that approximately covers the parameter trajectory. Instead of handling the continuous trajectory, we practically use its discretization, i.e., discretely sampled points, to characterize the trajectory. 
The basic operations include:
\begin{itemize}
    \item First, sample $t$ steps of neural network parameters during the training, namely,   $\{\mathbf{w}_1,\mathbf{w}_2,\ldots,\mathbf{w}_t\}$.

\item Second, centralize these samples as $\overline{\mathbf{w}}=\frac{1}{t}\sum_{i=1}^t \mathbf{w}_i$ and $W=[\mathbf{w}_1-\overline{\mathbf{w}},\mathbf{w}_2-\overline{\mathbf{w}},\ldots,\mathbf{w}_t-\overline{\mathbf{w}}]$.

\item Third, find a $d$-dimensional subspace spanned by $P=[\mathbf{e}_1, \mathbf{e}_2, \ldots, \mathbf{e}_d]$ to cover $W$. Notice that in DNNs the number of parameters $n$ is commonly significantly larger than $t$ and $d$.
\end{itemize}

The third step is to find a subspace that the distance of $W$ and the projection $P^\top W$ is minimized. With the ${l}_2$ norm, it could be formulated as  maximizing the variance of projection of $W$, i.e.,
\begin{equation}\label{PCA}
\begin{split}
&\max _{P} \quad \mathrm{tr} \left(P^{\top} W W^{\top} P\right),
  \\
&\quad \mathrm{s.t.} \quad P^{\top}P=I.
\end{split}
\end{equation}
This is a standard PCA problem that can be solved by performing spectral decomposition on $WW^{\top}$. The eigenvectors corresponding to the largest $d$ eigenvalues are orthonormal bases, or equivalently, the independent variables that we want for learning.

However, $WW^{\top}$ is an $n \times n$ matrix which has difficulties in storing, not to mention the high cost of its spectral decomposition.
Notice that $WW^{\top}$ is low-rank since $n$ is far greater than $t$. We alternatively consider the SVD of $W$:
\begin{equation}
    W=U\Sigma V^{\top},
\end{equation}
where $U=[\mathbf{u}_1,\mathbf{u}_2,\ldots,\mathbf{u}_n]$, $\Sigma=\mathrm{diag}(\sigma_1, \sigma_2,\ldots,\sigma_t) $, and $V=[\mathbf{v}_1,\mathbf{v}_2,\ldots, \mathbf{v}_t]$. The first $d$ columns of $U$ are the independent variables that we want herein.
Since $W$ and $W^{\top}$ essentially share the same SVD decomposition, we can first compute $\mathbf{v}_i, i=1, \ldots, d$ by the spectral decomposition of $W^{\top}W$, which is only a  squared matrix of $t\times t$, so that the vectors  $\mathbf{u}_i$ can be computed as
\begin{equation}
 W \mathbf{v}_i = \sigma_i \mathbf{u}_i, \ i=1, \ldots, d.
\end{equation}
In summary, our dimensionality reduction algorithm, i.e., DLDR, is given in Algorithm \ref{DLDR}.

\begin{algorithm}[h]
\caption{Dynamic Linear Dimensionality Reduction (DLDR)}
\label{DLDR}
\begin{algorithmic}[1]
\State Sample  parameter trajectory  $ \{\mathbf{w}_1,\mathbf{w}_2,\dots,\mathbf{w}_t\}$ along the training;
\State $\overline{\mathbf{w}}=\frac{1}{t}\sum_{i=1}^t \mathbf{w}_i$;
\State $W=[\mathbf{w}_1-\overline{\mathbf{w}},\mathbf{w}_2-\overline{\mathbf{w}},\ldots,\mathbf{w}_t-\overline{\mathbf{w}}]$;
\State Perform spectral decomposition on $W^{\top}W$ and obtain the largest $d$ eigenvalues $[\sigma_1^2,\sigma_2^2,\dots,\sigma_d^2]$ with the corresponding eigenvectors
$[\mathbf{v}_1,\mathbf{v}_2,\dots,\mathbf{v}_d]$;
\State $ \mathbf{u}_i = \frac{1}{\sigma_i} W \mathbf{v}_i$;
\State Return $[\mathbf{u}_1,\mathbf{u}_2,\dots,\mathbf{u}_d]$ as the orthonormal bases.
\end{algorithmic}
\end{algorithm}

For complexity, DLDR involves a spectral decomposition of a $t\times t$ matrix and two matrix productions. The total time complexity is $\mathcal{O}(t^3+t^2n+t^2n)$ and the matrix operations involved in DLDR can be greatly sped up by GPUs. Generally, the time consumption is negligible compared to DNNs training.

\subsection{Training Performance}
\label{sec3.3}
The proposed DLDR can reduce the dimensionality of  optimization space from $n$ to $d$, based on the hypothesis that the optimization trajectory approximately lies in a low-dimensional subspace.
To verify the hypothesis, we optimize DNNs in such low-dimensional subspaces, checking whether the performance could be similar to training over all parameters in the original space.

Firstly, we conduct experiments on training ResNet8 \cite{he2016deep} for CIFAR-10, which is also considered by 
the pioneering work \cite{gressmann2020improving}. As comparison/judgment, we use SGD to train 78330 parameters of ResNet8 from a random initialization. The detailed setting is: learning rate is 0.1, batch size is 128. Notice that in the experiments of this paper, the SGD always contains momentum term and here the momentum parameter is 0.9. With 3 trials, SGD gets averagely $83.84\%$ accuracy on test set.

\begin{figure}[!t]
	\centering
	\subfigure[]{ \includegraphics[angle=0,origin=br,width=6cm]{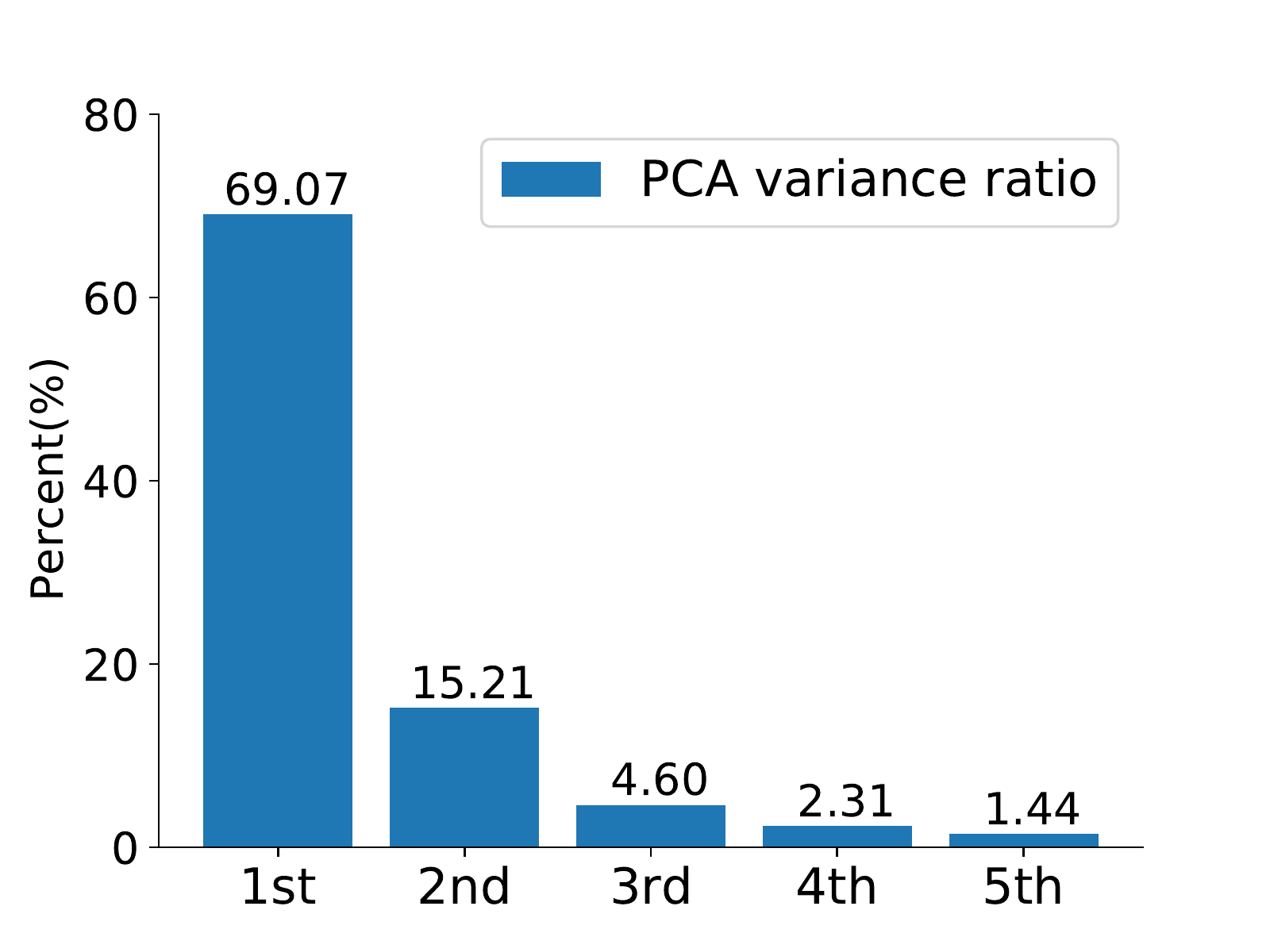}
    \label{traj_2:a}}\\
	\subfigure[]{ \includegraphics[angle=0,origin=br,width=6cm]{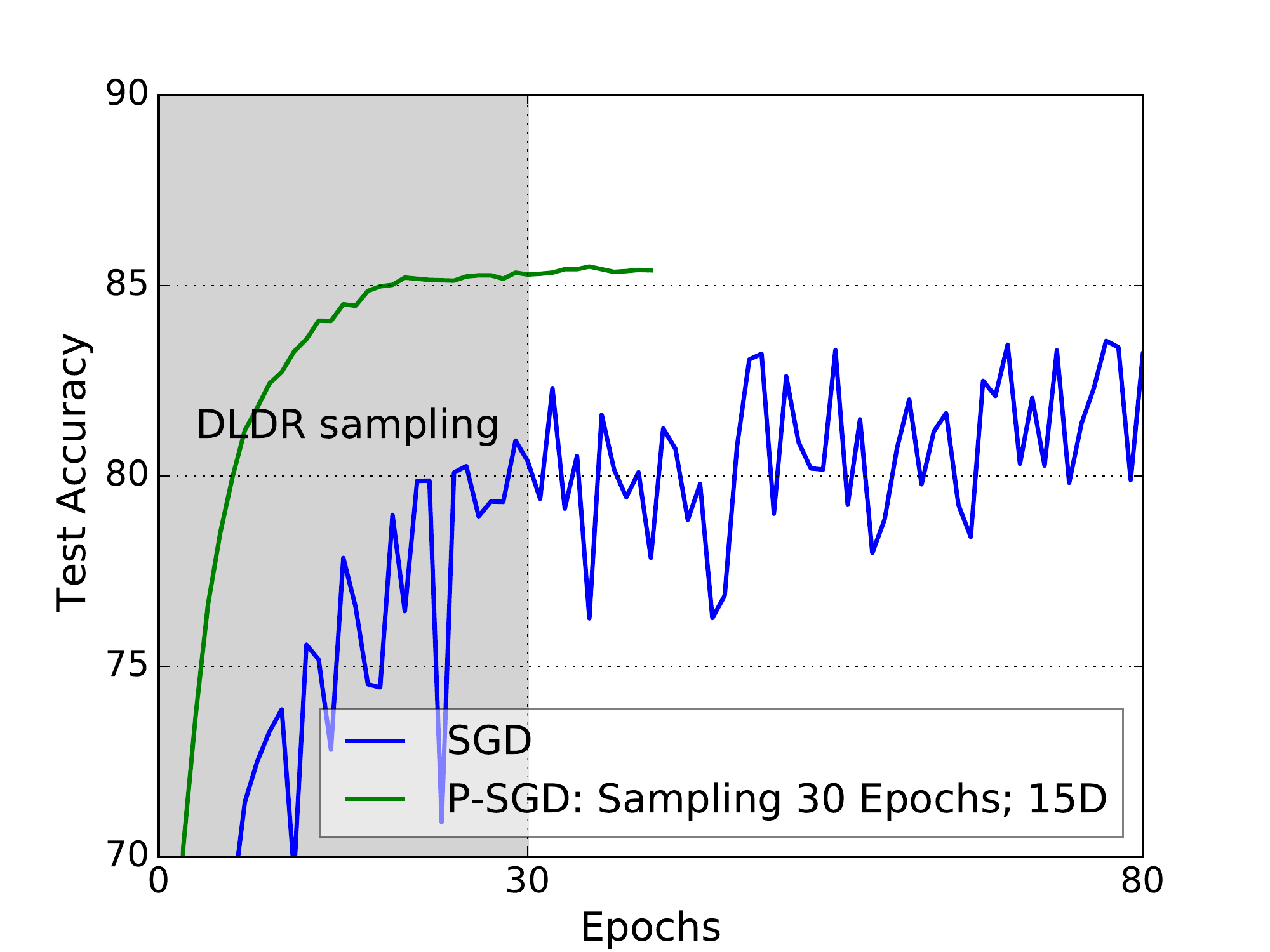}
    \label{traj_2:b}}
	\caption{Experiments for ResNet8 trained on CIFAR-10. (a) the PCA ratio of the first 30 epochs training trajectory in the principal directions. (b) the training performance of SGD and P-SGD (in 15D subspace).}
	\label{traj_2}
\end{figure}

Then, we apply DLDR to extract the low-dimensional subspace from the parameters by sampling the trajectory over 30 epochs of SGD training. The detailed sampling strategy is that we sample the model parameters after every epoch training. In Fig. \ref{traj_2:a}, the variance ratios of the top 5 projected components are plotted, showing that over 90\% of the total variance is owned to these five components.
This observation coincides with our hypothesis on the existence of such a low-dimensional subspace that can approximately cover the optimization trajectory.

Next, in the subspace extracting by DLDR, we train the neural networks from \emph{scratch}. The dimension we set here is 15 and
we use the SGD optimizer in the projected subspace, named as \emph{P-SGD}. To avoid other affects for fairness, we use the same hyper-parameter setting as regular SGD and start from the same initialization. From Fig. \ref{traj_2:b}, it can be seen that P-SGD quickly surpasses the performance when the DLDR sampling stops (therefore the good performance of P-SGD is not from the passable solution given by the DLDR sampling stage) and reaches an accuracy similar to or even better than regular SGD. Here, the superiority over SGD may come from de-noising (variance reduction) effect of low-dimensional subspace and could be further investigated. Yet at least, it shows that we can effectively train ResNet8 for CIFAR-10 in a subspace with a significantly lower dimensionality, i.e., 15,
which strongly supports our hypothesis that the optimization trajectory can approximately lie in a low-dimensional subspace.

In Table \ref{previousworks}, we report the dimension used for optimization and the test accuracy. In \cite{li2018measuring}, 7982 dimensions were used to achieve averagely $58.35\%$ accuracy, which was improved to $70.26$ in \cite{gressmann2020improving}. Now with training dynamics information, DLDR can find a subspace with much fewer dimensions and achieve significantly better accuracy. In Section 5, we will consider more complicated DNN architectures and more complicated tasks to further verify the low-dimensional subspace hypothesis.


\begin{table}[!ht]
	\centering
	\caption{Performance of training ResNet8 in low-dimensional subspaces for CIFAR-10 (optimizer: SGD)}
	{
	\footnotesize
		\begin{tabular}{lrr} 
    \toprule
		dimension reduction method & dimension    & test accuracy\\
	\midrule
		---  &78330 &83.84$\pm$0.34 \\
	\midrule
	Li \textit{et al.} \cite{li2018measuring} &7982 &58.35$\pm$0.04\\
	Gressmann \textit{et al.} \cite{gressmann2020improving} &7982 &70.26$\pm$0.02\\
	DLDR (ours) &\textbf{15} & \textbf{85.18$\pm$0.20} \\
	\bottomrule
	\end{tabular}
	}
	\label{previousworks}

\end{table}

\section{DLDR-based Quasi-Newton Algorithm}
\label{BFGSsec}
Since DNNs generally have massive parameters to optimize, first-order methods, i.e., gradient-descent-based methods, are the dominating methodology. However, there are some fundamental limitations in first-order methods, such as slow convergence around optima and high sensitivity to the learning rate. For these problems, second-order methods could be the remedy, but due to the high computational burden, 
there are great difficulties to apply them in training DNNs involving massive parameters. Instead, part of second-order information, like momentum and accumulation information, has been used, resulting in many popular training algorithms, like Adam \cite{kingma2014adam}, RMSprop \cite{dauphin2015rmsprop}, and AdaGrad \cite{mcmahan2010adaptive, duchi2011adaptive}.  Now with the proposed DLDR, it becomes possible to find only a few (dozens of) independent variables to optimize, which makes it applicable to use second-order methods in training complex DNNs. Following this idea, we develop a quasi-Newton method based on the framework of BFGS \cite{broyden1970convergence,byrd1994representations}. Analogously, the main steps include Hessian matrix approximation based on historical gradients, quasi-Newton update, and backtracking line search, of which the details are given in the following subsections.

\subsection{Hessian Matrix Approximation}
In Newton method, the descent direction is given as
\[
\mathbf{q} = -H^{-1} \vg,
\]
where $\vg$ is the gradient and $H \in \reals^{n \times n}$ is the Hessian matrix. When $n$ is large, it is computationally intractable to calculate the inverse of $H$, unless we can well approximate it as
\[
H \approx P H_0 P^\top,
\]
where $H_0$ is the Hessian matrix in the subspace
with orthonormal bases $\mathbf{e}_1,\mathbf{e}_2,\ldots,\mathbf{e}_d$, i.e., $H_0 \in \reals^{d \times d}$. With a small $d$, its inverse could be efficiently calculated,
and then the Newton direction  becomes
\begin{equation}\label{newton-direction}
    \mathbf{q}=-H^{-1} \mathbf{g}\approx -(P H_0 P^\top)^{\dagger}\mathbf{g}=-P H_0^{-1}P^\top \mathbf{g},
\end{equation}
where ${\dagger}$ denotes the pseudo-inverse operator.

For DNNs, the low-dimensional hypothesis indicates that we can do optimization in a tiny subspace with dimension $d$, where the Newton direction is given by (\ref{newton-direction}). Specifically,
the procedure is given as follows:
\begin{enumerate}
    \item $P^\top\vg$: project the gradient of parameters to the independent variable space;
    \item $-H_0^{-1}(P^\top\vg)$: calculate the Newton direction in the independent variable space;
    \item  $P (-H_0^{-1}(P^\top\vg))$: back project the Newton direction to the original parameter space, while the projection matrix is kept the same during the training.
\end{enumerate}

\subsection{Quasi-Newton Update}
Although we can find only a few independent variables to optimize, their gradients are calculated by projecting gradients of the original parameters in the current DNN framework. Therefore, directly calculating  the second-order gradient is still impractical. Alternatively, we adopt quasi-Newton method to approximate the Hessian matrix and its inverse. In this way, the standard BFGS algorithm \cite{broyden1970convergence, byrd1994representations} is used with the rank-two correction update as follows,
\begin{equation}
\begin{split}
    B_{k+1}&=V^{\top}_k B_k V_k + \rho_k \tilde{\mathbf{s}}_k \tilde{\mathbf{s}}_k^{\top}, \\
\mathbf{y}_k&=\tilde{\mathbf{g}}_{k+1}-\tilde{\mathbf{g}}_{k},\\
\rho_k&=(\mathbf{y}_k^{\top}\tilde{\mathbf{s}}_k)^{-1},\\
V_k&=I-\rho_k\mathbf{y}_k \tilde{\mathbf{s}}_k^{\top},
\end{split}
\label{updateBK}
\end{equation}
where $B_k$ is the inverse Hessian approximation matrix in the $k$-th step, $\tilde{\mathbf{g}}_{k}$ is the projected gradient by $P^\top \mathbf{g}_k$, and $\tilde{\mathbf{s}}_k = P^\top(\mathbf{w}_{k+1} -\mathbf{w}_k)$ is the projected difference between the parameters in the successive two steps. Here, we begin with initializing the inverse Hessian approximation  as $B_0=I$.

Notice that the above BFGS algorithm is always working in the subspace, so the involved matrices and vectors are all related to dimension $d$, instead of $n$. Therefore, the computational complexity is very low
and it is totally different to the existing BFGS-based training methods \cite{liu1989limited, yuan1991modified, mokhtari2014res, botev2017practical, bollapragada2018progressive,goldfarb2020practical} that are working in the original $n$-dimensional space.

\subsection{Backtracking Line Search}
DNNs are highly non-convex and  the Hessian matrix
is not always positive semi-definite thereby. To ensure the loss descent, BFGS  requires $\mathbf{y}_k^{\top}\tilde{\mathbf{s}}_k>0$,  guaranteeing the positiveness of the Hessian approximation $B_{k+1}$.
In the proposed method, we adopt the backtracking line search to satisfy the Armijo condition \cite{armijo1966minimization},
\begin{equation}
\label{linesearch}
\begin{split}
    \mathcal{L}_{\mathcal{B}_k}(\mathbf{w}_k - \alpha_k P B_k \tilde{\mathbf{g}}_k)~~~~~~~~~~~~~~ \\
    \le  \mathcal{L}_{\mathcal{B}_k}(\mathbf{w}_k )-c \alpha_k \tilde{\mathbf{g}}_k^{\top} B_k \tilde{\mathbf{g}}_k,
\end{split}
\end{equation}
where $\mathcal{L}_{\mathcal{B}_k}$ is the loss with the mini-batch  $\mathcal{B}_k$, and $c$ is a positive constant. We start from $\alpha_k=1$ and repeatedly multiply it by a constant factor $\beta \in (0,1)$ until \eqref{linesearch} holds true. In this paper, we adopt $c=0.4$ and $\beta=0.55$, which empirically shows good performance and is of course not necessarily optimal for all scenarios.

\subsection{Algorithm Summary}
We now summarize the developed quasi-Newton algorithm based on DLDR in
Algorithm \ref{P-BFGS}. Since it is essentially a BFGS in the projected subspace, we name it as \emph{P-BFGS}.

\begin{algorithm}
\caption{P-BFGS}\label{P-BFGS}
\begin{algorithmic}[1]
\State Obtain $P = [\mathbf{e}_1,\mathbf{e}_2,\dots,\mathbf{e}_d]$
by DLDR in Algorithm \ref{DLDR};
\State Initialize $k \gets 0$ and $B_k\gets I$;
\While{not converging}
\State Sample the mini-batch data $\mathcal{B}_k$;
\State Compute the  gradient $\mathbf{g}_k$ on $\mathcal{B}_k$;
\State Perform the projection $\tilde{\mathbf{g}}_k \gets P^{\top}\mathbf{g}_k$;
\If {$k>0$}
\State Do Quasi-Newton update $B_k$ with  \eqref{updateBK};
\EndIf
\State Compute $\alpha_k$ using backtracking line search;
\State $\tilde{\mathbf{s}}_k \gets -\alpha_k B_k \tilde{\mathbf{g}}_k$;
\State $\mathbf{w}_{k+1} \gets \mathbf{w}_k + P \tilde{\mathbf{s}}_k$; \Comment{Update the parameters}
\State $k \gets k + 1$;
\EndWhile
\end{algorithmic}
\end{algorithm}


\section{Numerical Experiments}
\label{Experiments}
In this section, numerical experiments are conducted on different tasks and different neural network architectures. First, we apply P-SGD to train DNNs in the subspaces, of which the dimension is fixed to {\bf 40}, extracted by DLDR. If the training performance is similar to training over full parameters in the original space, we can verify the low-dimensional landscape hypothesis. Second, based on the low-dimensional property,  we  evaluate the performance of the proposed P-BFGS algorithm. Notice that our purpose is not to claim the advantage of P-BFGS over other training methods, but to use the feasibility of second-order methods to further support our main claim: one can train DNNs in tiny subspaces. Third, we conduct experiments with label noises to show the inherent robustness benefiting from training in the obtained low-dimensional subspaces.

\subsection{Experiments Setup}
The datasets used in our experiments include CIFAR-10, CIFAR-100 \cite{krizhevsky2009learning}, and ImageNet \cite{deng2009imagenet}.
For CIFAR, all images are normalized by channel-wise mean and variance. Data augmentations \cite{he2016deep} are also performed: horizontal image flipping with probability 0.5, 4-pixel padding, and cropping.
We test on ResNet20 and ResNet32 \cite{he2016deep}, and also other 11 DNN architectures.
The numbers of full parameters in these networks are from 0.27M to 28.5M, but we always choose only {\bf{40}} independent variables in this paper.
We train the DNNs using SGD \cite{ruder2016overview} and Adam \cite{kingma2014adam} optimizers, for which the weight decay is set as 1e-4, momentum parameter as 0.9, and batch size as 128. The default initial learning rates are set as 0.1 and 0.001 for SGD and Adam, respectively. For CIFAR-10, we train the DNNs for 150 epochs and divide the learning rate by 10  at 100 epochs while for CIFAR-100 we train 200 epochs and divide at 150 epochs.
For ImageNet, our code is modified from the official PyTorch example\footnote{\url{https://github.com/pytorch/examples/tree/master/imagenet}}. The experiments are performed on Nvidia Geforce GTX 2080 TI.
We use one GPU for CIFAR and four GPUs for ImageNet. Mean and standard deviation are obtained from 5 independent experiments.


DLDR needs to sample the optimization trajectory. 
For CIFAR, we adopt the simplest sampling strategy: the model parameters are sampled after every epoch of training.
For ImageNet, the parameters are uniformly sampled 3 times in each epoch of training. 
A more delicate sampling strategy may
improve the performance.

When training DNNs in the subspace by P-SGD,
we adopt the same batch size and momentum factor as SGD. We set the initial learning rate as 1, training epoch as 40 and divide the learning rate by 10 at 30 epochs.
For P-BFGS, we set batch size as 1024 for CIFAR and 256 for ImageNet. As a second-order algorithm, it does not need a learning rate schedule.

\subsection{Verification on Various Architectures}
In subsection \ref{sec3.3}, experiments are conducted on CIFAR-10 and here we verify the low-dimensional landscape hypothesis on CIFAR-100.
We will train DNNs by SGD on all the parameters and in the reduced subspace (the latter actually is the proposed P-SGD), respectively. For different neural network architectures, we always choose 40 independent variables.
If SGD and P-SGD give comparable performance, it supports our hypothesis and meanwhile verifies the effectiveness of the proposed DLDR.

This experiment contains 11 popular DNNs, including VGG11 \cite{simonyan2014very}, DenseNet121 \cite{huang2017densely}, Inception \cite{szegedy2016rethinking}, NasNet\cite{zoph2018learning}, etc., and the number of parameters varies from 780K to 28.5M. In TABLE \ref{cifar100variosus}, we report the test accuracy using SGD with 50/100/200 epochs and the test accuracy after 200 epochs serves as the baseline. We then apply P-SGD in 40D subspaces, which are extracted by DLDR from 50 or 100 epochs sampling, and conduct 40 epochs training from the initialization.
The results are reported in the last column in TABLE \ref{cifar100variosus}, which clearly shows that P-SGD with 40 independent variables could reach competitive performance of SGD over all parameters. This competitive performance is verified in all these architectures and thus strongly supports our low-dimensional landscape hypothesis. Minor finding is that generally the performance would be better if the subspaces are better extracted.


\begin{table*}[bhtp]
\begin{center}
\caption{Test accuracy on CIFAR-100 for regular training and optimization in 40D subspaces 
}\label{cifar100variosus}
\setlength{\belowcaptionskip}{-0.5cm}
{\footnotesize
\begin{tabular}{c|r|ccc|cc}
\toprule
\multirow{2}{*}{Models}
&\multirow{2}{*}{\# Parameter}
& \multicolumn{3}{c|}{SGD (\#training epochs) }
&\multicolumn{2}{c}{P-SGD (\#sample epochs) }\\
& & 50 &100 &200  & 50 & 100\\
\midrule
VGG11\_bn \cite{simonyan2014very} &28.5M &58.38 &59.90 &68.87 &68.72  &{70.18} \\
EfficientNet-B0 \cite{tan2019efficientnet} &4.14M &62.53 &63.68 &{72.94} &71.68 &72.64 \\
MobileNet \cite{howard2017mobilenets} &3.3M &57.15 &58.67  &{67.94} &66.86  &68.00 \\
DenseNet121 \cite{huang2017densely} &7.0M &65.39 &64.57 &{76.76} &74.25 &76.34\\ 
Inceptionv3 \cite{szegedy2016rethinking} &22.3M &61.68 &64.00 &76.25 &75.15  &{76.83}\\
Xception \cite{chollet2017xception}	&21.0M &64.57 &65.81	&75.47	&{75.68}		&75.56\\
GoogLeNet \cite{szegedy2015going}	&6.2M &62.32 &66.32	&76.88		&75.66		&{77.27} \\
ShuffleNetv2 \cite{ma2018shufflenet}	&1.3M &62.90 &63.15	&72.06	&71.34		&{72.29}\\
SequeezeNet \cite{iandola2016squeezenet} &0.78M &59.52 &58.56 &70.29 &69.89 &{70.60} \\
SEResNet18 \cite{hu2018squeeze} &11.4M &64.74 &64.68  &74.95 &74.33  &{75.09} \\
NasNet \cite{zoph2018learning} &5.2M &63.73 &66.80 &{77.34} &77.19 &77.03  \\
\bottomrule
\end{tabular}
}
\end{center}
\end{table*}

\subsection{Training Performance of P-BFGS Algorithm}
\label{Exp_CIFAR_10}
\begin{figure}[htbp]
	\centering
	\subfigure[]{ \includegraphics [width = 7 cm]{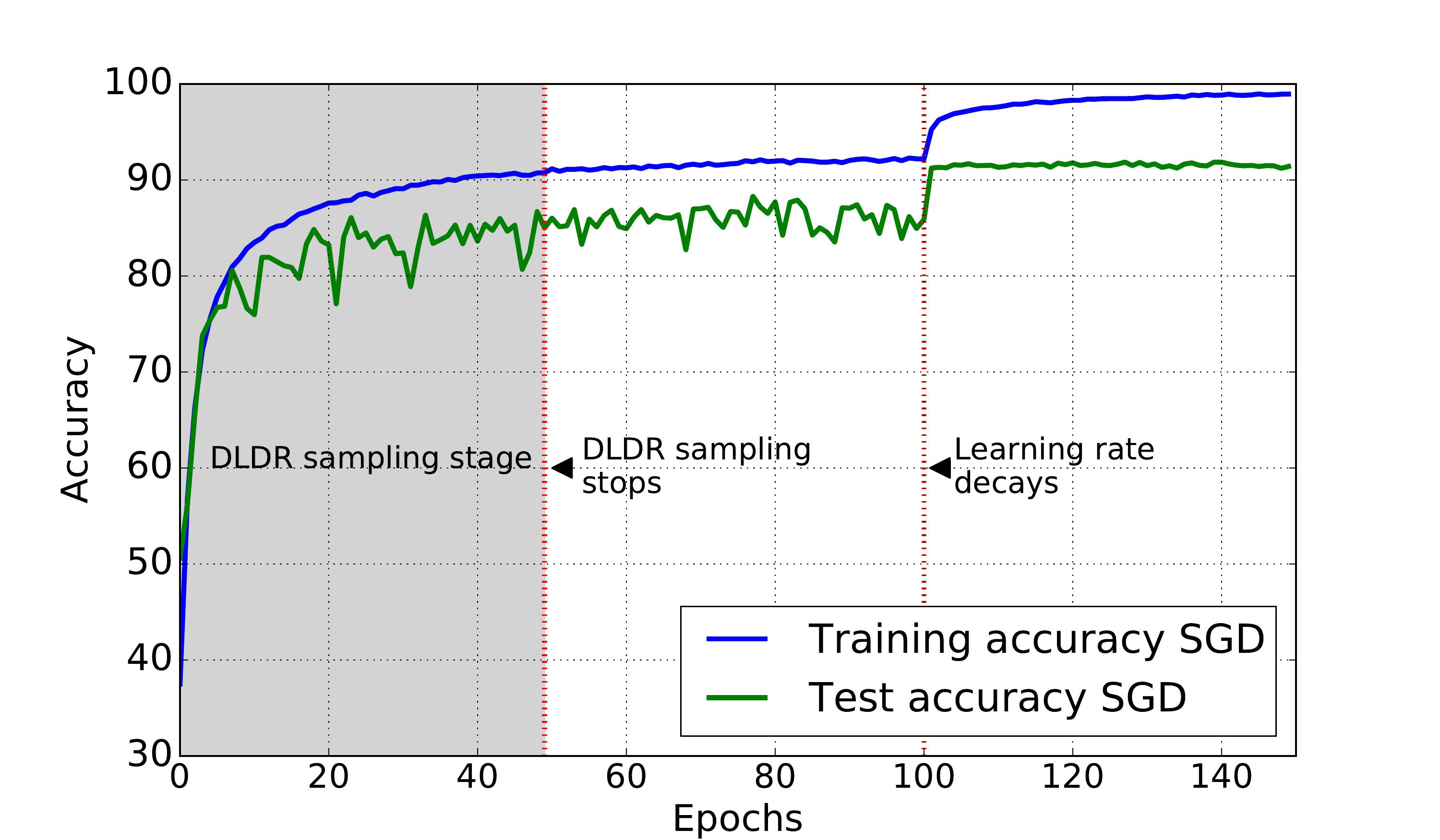}
    \label{cifar10SGD:a}}

	\subfigure[]{ \includegraphics [width = 7 cm]{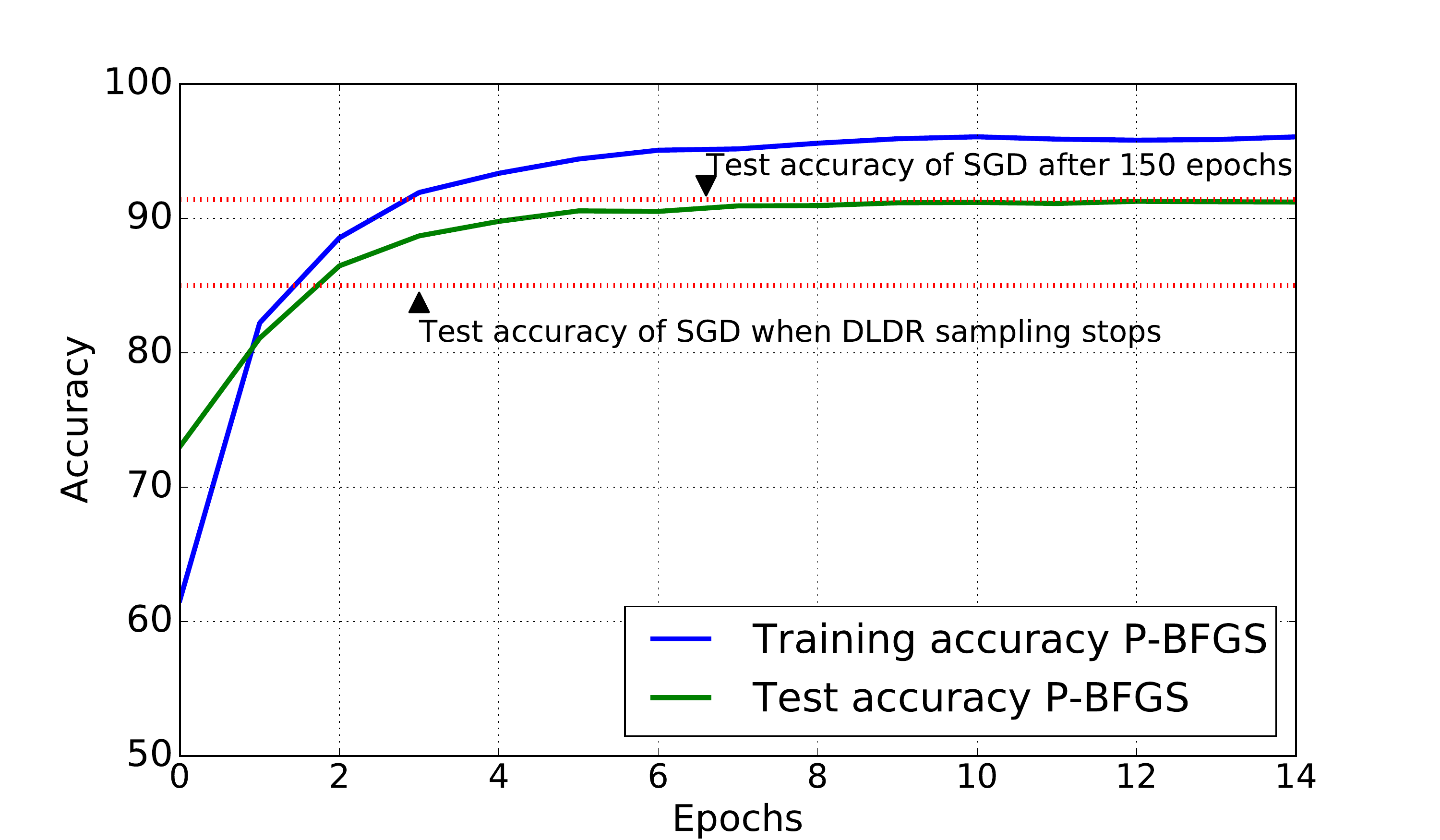}
    \label{cifar10SGD:b}}
    \caption{The accuracy of ResNet20 on CIFAR-10 in different epochs. (a)  SGD on 0.27M parameters: DLDR samples the first 50 epochs and there is learning rate decay after 100 epochs. (b) P-BFGS on 40 variables: trains from the initial, P-BFGS could surpass the DLDR sampling stage very quickly and achieve similar performance as (a). }
	\label{cifar10SGD}
\end{figure}

After empirically demonstrating the low-dimensional landscape hypothesis, we now try second-order algorithm, namely P-BFGS.
We firstly consider ResNet20 \cite{he2016deep} on CIFAR-10. In Fig. \ref{cifar10SGD:a}, the training and test accuracy curves of SGD are plotted. The gray region indicates where we get samples for DLDR and then extract the independent variables. After obtaining the $40$ independent variables, we use P-BFGS starting from the same initialization and plot the training curves in Fig. \ref{cifar10SGD:b}.
After only 2 epochs, P-BFGS attains better performance than SGD with 50 epochs, i.e., the samplings for DLDR, and within 10 epochs, P-BFGS arrives at the performance of SGD with 150 epochs, which preliminarily demonstrates the advantages of applying second-order methods in efficiency.

\begin{figure}[!t]
\centering
\includegraphics[width=0.9\linewidth]{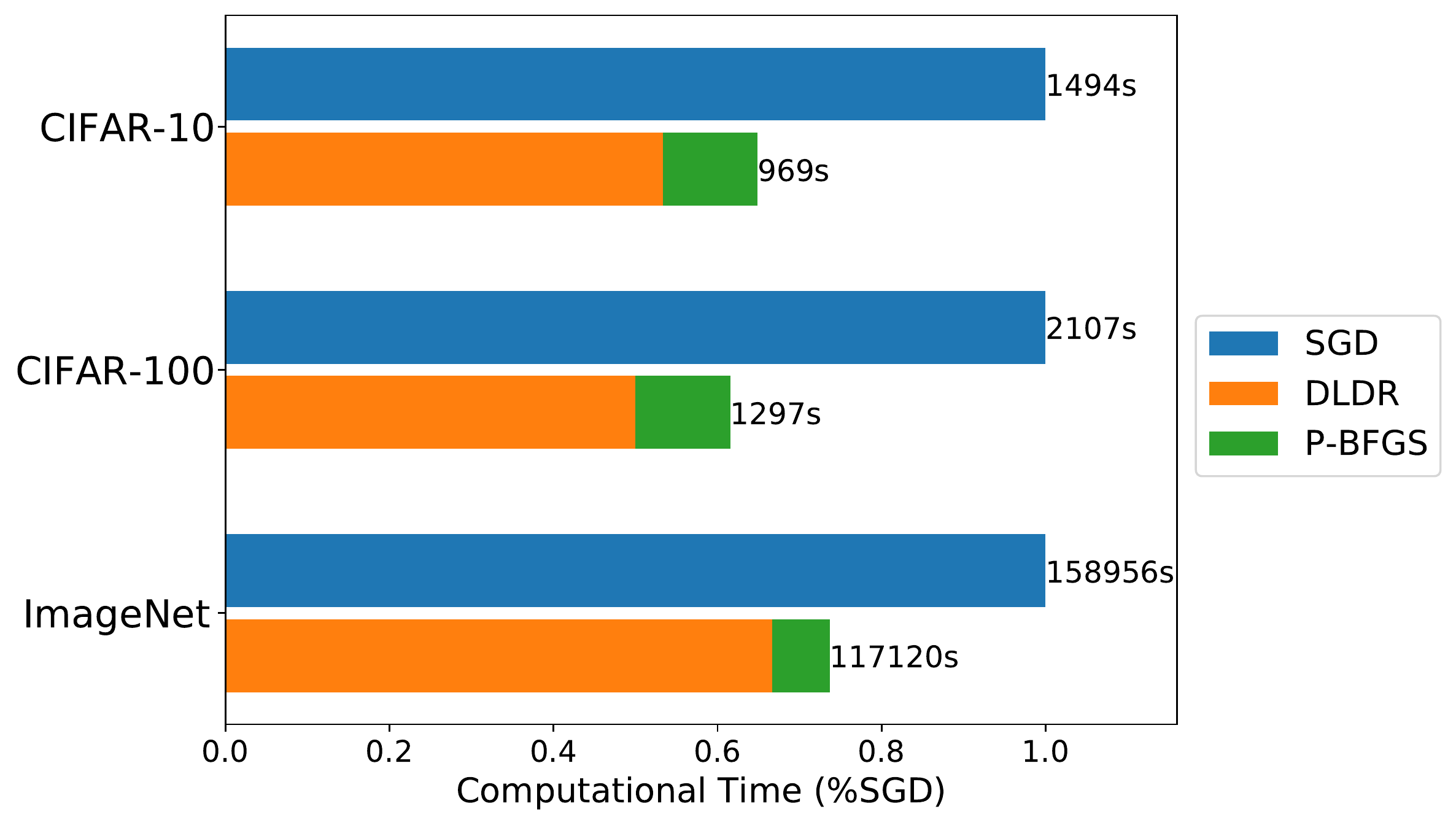}
\caption{Wall-clock time consumption comparisons.}
\label{time_consumption}
\end{figure}

Next, we report quantitative comparisons between SGD and P-BFGS on both accuracy and computational time. The tasks include CIFAR-10, CIFAR-100, and ImageNet. For CIFAR-10 and CIFAR-100, we extract 40D subspaces from 80 and 100 epochs of SGD, respectively. ImageNet is a more challenging task and requires more independent variables, for which 120 independent variables are obtained from 60 epochs of SGD.  In TABLE \ref{pbfgs_table-simple}, we report the test accuracy comparisons. Generally, the accuracy of P-BFGS could be similar to that of SGD, showing again it is sufficient to optimize a DNN in a very low-dimensional space. Thanks to the fact that the number of optimization variables is small now, it is possible to use second-order methods, which may have benefits on optimization, e.g., fast convergence and getting riding of manually tuning learning rate.
The detailed wall-clock time comparisons are presented in Fig. \ref{time_consumption}, where we normalize the time according to the total training time of SGD.
For P-BFGS, the time consumption includes two parts: i) DLDR sampling; and ii) optimization in the subspaces.
As expected, applying second order methods could significantly improve the convergence speed: the epochs that P-BFGS requires are quite a few. Overall, P-BFGS can save around $30\%$ time from SGD. Notably, over $80\%$ of the total time is used in DLDR sampling. In the future, more sophisticated techniques of identifying the independent variables are promising to further speed up the training.

\begin{table}[htbp]
    \centering
    \caption{Test accuracy obtained by SGD and P-BFGS
    }
    \label{pbfgs_table-simple}
    {
    \footnotesize
    \begin{tabular}{c|c|c|c|c}
    \toprule
    \multicolumn{2}{c|}{\textbf{Dataset}} &\textbf{CIFAR-10} &\textbf{CIFAR-100} &\textbf{ImageNet} \\
    \midrule
    \multicolumn{2}{c|}{Model} &ResNet20 &ResNet32 &ResNet18\\
    \midrule
    \multirow{2}{*}{SGD} &epochs &150 &200 &90 \\
    &acc &$91.55 \pm 0.23$ &$68.40\pm 0.45$ &69.794\\
    \midrule
    \multirow{3}{*}{P-BFGS} &sampling &80 &100 &60 \\
    &epochs &20 &20 &4 \\
    &acc &$91.72 \pm 0.10$ &$69.90 \pm 0.48$ &69.720\\
    \bottomrule
    \end{tabular}
    }

\end{table}

\subsection{Robustness under label noises} \label{LabelNoise_sec}
Due to the interpolation essence of DNNs, they are very sensitive to label noises, i.e., when the labels are incorrect, DNNs will follow these incorrect and meaningless labels. Even worse, there is no explicit difference to distinguish whether do DNNs learn correct or incorrect labels \cite{zhang2021understanding}. Currently, only early stop  can be used \cite{li2020gradient} but how to choose the best stop is very challenging, since even validating data are also corrupted. Now we have certificated that DNNs can be trained in low-dimensional subspaces and expect that the low-dimensional property could naturally bring robustness against label noises.

\begin{figure}[!t]
\centering
\includegraphics[width=0.8\linewidth]{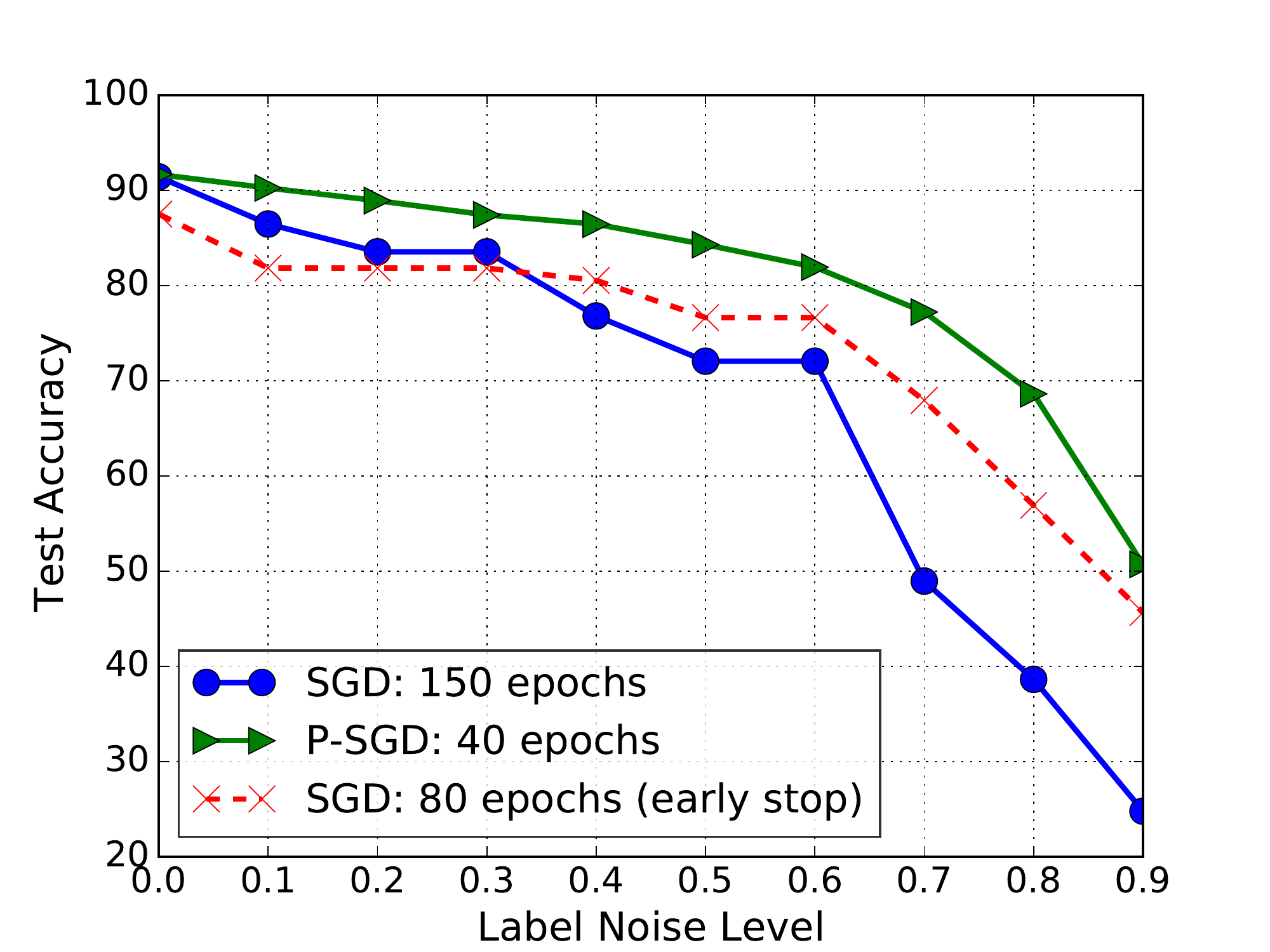}
\caption{The performance under different level of label noise.}
\label{label_noise_ratio}
\end{figure}

To examine the performance under label noises, we consider CIFAR-10 and randomly select a fraction $c$ of the training data and assign random labels to them (which are fixed for different methods). With label noise, the full training performance of SGD is significantly dropped, as plotted by a blue curve in Fig. \ref{label_noise_ratio}. Early stop (red curve) indeed can help but have less accuracy for clean data. Training in low dimensional subspaces (green curve) can consistently outperform the early stop with a large margin, while keep the same performance on clean data as regular training. Note that here we obtain robustness without any enhancement techniques, e.g., modifications on the loss function \cite{ghosh2017robust, zhang2018generalized}, and thus the results are promising to be further improved via these techniques.

The robustness of P-SGD comes from the low-dimensional property of the subspace, or in other words, the degree-of-freedom is quite small. To further investigate the effect of the number of independent variables, we vary $d$ from 10 to 40 and report the test accuracy in TABLE \ref{independent_variables_number}. With different noise level, the P-SGD with full training (i.e., we do not do early stop if the training loss is decreased) can always have better accuracy than SGD with full training. We also provide the best accuracy obtained by SGD, i.e., we monitor the test accuracy during the training and select the best test accuracy occurred.
The SGD (best) of course cannot be reached in practice but it can be used as a reference showing that training DNNs in the low-dimensional subspace is indeed robust to label noises.

\begin{table}[!ht]
	\centering
	\caption{The effects of independent variables' number}
\label{independent_variables_number}
{
\footnotesize
	\begin{tabular}{c|cccc|cc} 
    \toprule
		Noise  &\multicolumn{4}{c|}{P-SGD Final}  &SGD &SGD\\
		Level &$d=10$ &$d=20$ &$d=30$ &$d=40$  &Final &Best\\
	\midrule
	$c=0.7$ &68.92  &77.00 &77.01 &\textbf{77.22} &49.0 &73.4 \\
    $c=0.8$  &64.95  &\textbf{68.63} &68.51 &68.57 &38.6 &63.6\\
    $c=0.9$ &50.63  &50.49 &\textbf{51.00} &50.38 &24.8 &49.3\\
	\bottomrule
	\end{tabular}
}
\end{table}

\section{Conclusions and Further Works}
\label{discussion}
The starting point of this paper is the low-dimensional landscape hypothesis that DNNs' landscape function can be covered in a tiny subspace. Based on the training dynamic, we design an efficient dimension reduction method called DLDR. In comprehensive experiments, optimizing quite a few, e.g., dozens of, independent variables extracted by DLDR can attain similar performance as regular training over all parameters. Both the optimization and the reduction performance has been dramatically improved from the pioneer works \cite{li2018measuring, gressmann2020improving}, strongly supporting the hypothesis that DNNs can be trained in a tiny subspace.


From the new perspective of DNNs' training, two follow-up applications are tried out to further support our hypothesis and get great benefits from such low-dimensional property. First, as the dimensions substantially decrease, second-order methods become applicable, from which we design a P-BFGS algorithm and illustrate its  promising performance. Second, training in low-dimensional subspace naturally brings robustness against label noise, which is then verified by experiments.
These two applications  further support the low-dimensional subspace hypothesis. Although they are quite simple and straightforward, e.g., we simply use the original BFGS framework and do not apply any robust enhancement, their performance imply that finding the low-dimensional landscape could
benefit both theoretical and practice learning. Possible directions that follow the finding of low-dimensional subspace include to understand over-fitting, double descent \cite{nakkiran2019deep}, and over-parameterization \cite{allen2019convergence}, and to investigate few-shot learning, \cite{snell2017prototypical}, meta-learning \cite{vilalta2002perspective}, and transfer learning \cite{pan2009survey}, etc.

{\small
\bibliographystyle{ieee_fullname}
\bibliography{main}
}

\end{document}